\documentclass[conference]{IEEEtran}
\usepackage{cite}
\usepackage[pdftex]{graphicx}
\usepackage{subcaption}
\DeclareGraphicsExtensions{.pdf,.jpeg,.png}
\usepackage{amsmath}
\usepackage[linesnumbered,ruled,norelsize]{algorithm2e}
\makeatletter
\newcommand{\removelatexerror}{\let\@latex@error\@gobble}
\makeatother

\begin{document}

\title{Dynamic Path Planning and Replanning \\ for Mobile Robots using RRT*}

\author{\IEEEauthorblockN{Devin Connell}
\IEEEauthorblockA{Advanced Robotics and Automation Lab\\
Department of Computer Science and Engineering\\
University of Nevada, Reno NV, 89519\\
Email: devin.connell@gmail.com}
\and
\IEEEauthorblockN{Hung Manh La}
\IEEEauthorblockA{Advanced Robotics and Automation Lab\\
Department of Computer Science and Engineering\\
University of Nevada, Reno NV, 89519\\
Email: hla@unr.edu}}

\maketitle
\begin{abstract}
It is necessary for a mobile robot to be able to efficiently plan a path from its starting, or current, location to a desired goal location.  This is a trivial task when the environment is static.  However, the operational environment of the robot is rarely static, and it often has many moving obstacles.  The robot may encounter one, or many, of these unknown and unpredictable moving obstacles.  The robot will need to decide how to proceed when one of these obstacles is obstructing it's path.  A method of dynamic replanning using RRT* is presented.  The robot will modify it's current plan when an unknown random moving obstacle obstructs the path.  Various experimental results show the effectiveness of the proposed method.
\end{abstract}

\IEEEpeerreviewmaketitle

\section{Introduction}

    Path planning has been one of the most researched problems in the area of robotics.  The primary goal of any path planning algorithm is to provide a collision free path from a start state to an end state within the configuration space of the robot.  Probabilistic planning algorithms, such as the Probabilistic Roadmap Method (PRM) \cite{Kavraki_96} and the Rapidly-exploring Random Tree (RRT) \cite{Lavalle98}, provide a quick solution at the expense of optimality.  Since its introduction the RRT algorithm has been one of the most popular probabilistic planning algorithms.  The RRT is a fast, simple algorithm that incrementally generates a tree in the configuration space until the goal is found.    
    
    The RRT has a significant limitation in finding an asymptotically optimal path, and has been shown to never converge to an asymptotically optimal solution \cite{Karaman_2} \cite{Karaman_1}.  There is extensive research on the subject of improving the performance of the RRT.   Simple improvements such as the Bi-Directional RRT and the Rapidly-exploring Random Forest (RRF) improve the search coverage and speed at which a single-query solution is found.  The Anytime RRT \cite{Ferguson_1} provides a significant improvement in cost-based planning.  The RRT* algorithm provides a significant improvement in the optimality of the RRT and has been shown to provide an asymptotically sub-optimal solution \cite{Karaman_2}.

Since the introduction of the RRT* algorithm, research has expanded to discover new ways to improve upon the algorithm.  Research includes adding heuristics \cite{Perez_1} \cite{ Gammell_1} or bounds \cite{Salzman_1} to the algorithm in order to maintain the convergence of the algorithm but reduce the execution time.  Additional research attempts to guide the algorithm through intelligent sampling \cite{Islam_1}, or guided sampling through an artificial potential field \cite{Qureshi_1}.       

In many scenarios the operational environment is rarely static.  The path from a single query will often be obstructed during execution.  For that reason the topic of replanning is very important to robotic path planning.  It is not feasible to discard an entire search tree and start over.  One method is to store waypoints and regrow trees called the ERRT \cite{Bruce_2002}.  Another method (DRRT) is to place the root of the tree at the goal location, so that only a small number of branches may be lost or invalidated when replanning \cite{Ferguson_2}.   The Multipartite RRT maintains a set of subtrees that may be pruned and reconnected, along with previous states to guide regrowth.  It is essentially a combination of DRRT and ERRT \cite{Kuffner_1}.  More recently the RRT* algorithm has been incorporated into replanning.  RRT\textsuperscript{X} is an algorithm that uses RRT* to continuously update the path during execution \cite{Otte}.  The RRT\textsuperscript{X} is able to compensate for instantaneous changes in the static environment which is outside the scope of this work.

The contribution of this paper is the method using the RRT* algorithm for replanning in a dynamic environment with random, unpredictable moving obstacles.  Also included is the comparison of RRT and RRT* algorithms in a complex 2-D environment.  

The remainder of this paper is organized as follows:  Section II provides an overview of the RRT* algorithm.  Section III will present the replanning method using RRT* in a dynamic environment.  Section IV contains the results from all simulations.  Section V presents the conclusions and future work.

\section{Robot Path Planning using the RRT*}

The RRT* algorithm provides a significant improvement in the quality of the paths discovered in the configuration space over it's predecessor the RRT.  The quality of the path is determined by the cost associated with moving from the start location to the end location.  While RRT* does produce higher quality paths, the algorithm does have a longer execution time.  The longer execution time of RRT* is due to the algorithm making many additional calls to the local planner in order to continuously improve the discovered paths.  
  RRT* operates in a very similar way as RRT.  The algorithm builds a tree using random samples from the configuration space of the robot and connects new samples to the tree as they are discovered.  There are two primary differences between RRT and RRT*.  The first difference is in the method that new edges are added to the tree.  The second difference is an added step to change the tree in order to reduce path cost using the newly added vertex.  Each of these differences contributes to the improvement of discovered paths over time and is reason RRT* will converge to an asymptotically sub-optimal solution.

When a random vertex is added to the tree, the RRT will select the nearest neighbor as the parent for this new vertex and edge.  RRT* will select the best neighbor as the parent for the new vertex.  While finding the nearest neighbor, RRT* considers all the nodes within a neighborhood of the random sample.  RRT* will then examine the cost associated with connecting to each of these nodes.  The node yielding the lowest cost to reach the random sample will be selected as the parent, and the vertex and edge are added accordingly.  

\begin{figure}[!t]
\removelatexerror
\begin{algorithm}[H]
\DontPrintSemicolon
\caption{$T = (V, E) \leftarrow $RRT*($q_{init}$)}
$T \leftarrow $InitializeTree() \;
$T \leftarrow $InsertNode($\emptyset$,$q_{init}$,$T$) \;  

 \For{$k\leftarrow 1$ \KwTo $N$}{
  $q_{rand} \leftarrow $RandomSample($k$)\;
  $q_{nearest} \leftarrow $NearestNeighbor($q_{rand}$,$Q_{near}$,$T$)\;
  $q_{min} \leftarrow $ChooseParent($q_{rand}$,$Q_{near}$,$q_{nearest}$, $\Delta$q)\;
  $T \leftarrow $InsertNode($q_{min}$, $q_{rand}$, $T$) \;
  $T \leftarrow $Rewire($T$, $Q_{near}$, $q_{min}$, $q_{rand}$)\;
 }
\end{algorithm}
\vspace{-10pt}
\end{figure}

The RRT* algorithm begins in the same way as the RRT.  However, when selecting the nearest neighbor the algorithm also selects the set of nodes, $Q_{near}$, in the tree that are in the neighborhood of the random sample $q_{rand}$.  Line 6 of Algorithm 3 is the first major difference between RRT* and the RRT.  Instead of selecting the nearest neighbor to the random sample, the $ChooseParent()$ function will select the best parent from the neighborhood of nodes.

\begin{figure}[!t]
\removelatexerror
\begin{algorithm}[H]
\DontPrintSemicolon

\caption{$q_{min} \leftarrow $ChooseParent($q_{rand}$,$Q_{near}$,$q_{nearest}$,$\Delta$q)}
$q_{min} \leftarrow q_{nearest}$ \;
$c_{min} \leftarrow $Cost($q_{nearest}$) + c($q_{rand}$)\;
\For{$q_{near} \in Q_{near}$ }{
  $q_{path} \leftarrow $Steer($q_{near}$, $q_{rand}$, $\Delta$q)\;
\If{ObstacleFree($q_{path}$)} 
{
  $c_{new} \leftarrow $Cost($q_{near}$) + c($q_{path}$) \;
  \If{$c_{new}$ $<$ $c_{min}$}
  {
	$c_{min} \leftarrow c_{new}$ \;
	$q_{min} \leftarrow q_{near} $ \;
  }
}
}
\Return $q_{min}$ \;
\end{algorithm}
\vspace{-10pt}
\end{figure}

Algorithm 4 describes the $ChooseParent()$ function.  This function maintains the node with the lowest total cost for reaching $q_{rand}$.  At line 1 of Algorithm 4 the nearest neighbor, $q_{nearest}$, is considered the minimum cost neighbor, or $q_{min}$.  On line 2 the cost associated with reaching the new random sample $q_{rand}$ by using $q_{nearest}$ as the parent is stored as the current best cost, or $c_{min}$.  The algorithm then searches the set of nodes in the neighborhood of $q_{rand}$.  The $Steer()$ function on line 4 of Algorithm 4  will return a path from the nearby node, $q_{near}$ to $q_{rand}$.  If this path is obstacle free and has a lower cost than the current minimum cost, then the nearby node becomes the best neighbor, $q_{min}$ and this cost becomes the best cost $c_{min}$ (lines 7-9 of Algorithm 4).  When all nearby nodes have been examined the function returns the best neighbor.  The new random node is inserted into the tree using $q_{min}$ as the parent.  The next step is the second major difference between the RRT* and the RRT algorithms.  Line 8 of Algorithm 3 calls the $Rewire()$ function.

\begin{figure}[!t]
\removelatexerror
\begin{algorithm}[H]
\DontPrintSemicolon
\caption{$T \leftarrow $Rewire($T$, $Q_{near}$, $q_{min}$, $q_{rand}$)}
 \For{$q_{near} \in Q_{near}$ }{
	$q_{path} \leftarrow $Steer($q_{rand}$, $q_{near}$)\;
	\If{ObstacleFree($q_{path}$) and Cost($q_{rand}$) + c($q_{path}$) $<$ Cost($q_{near}$)} 
	{
		$T \leftarrow $ReConnect($q_{rand}$, $q_{near}$, $T$) \;
	}
 }
\Return $T$ \;

\end{algorithm}
\vspace{-21pt}
\end{figure}

The $Rewire()$ function, described in Algorithm 5, changes the tree structure based on the newly inserted node $q_{rand}$.  This function again uses the nearby neighborhood of nodes, $Q_{near}$, as candidates for rewiring.  The $Rewire()$ function uses the $Steer()$ function to get the path, except this time the path will start from the new node, $q_{rand}$ and go to the nearby node $q_{near}$.  If this path is obstacle free and the total cost of this path is lower than the current cost to reach $q_{near}$ (line 3 of Algorithm 5).  Then the new node $q_{rand}$ is a better parent than the current parent of $q_{near}$.  The tree is then rewired to remove the edge to the current parent of $q_{near}$, and add an edge to make $q_{rand}$ the parent of $q_{near}$.  This is done using the $ReConnect()$ function on line 4 of Algorithm 5.

The functions $ChooseParent()$ and $Rewire()$ change the structure of the search tree when compared to the RRT algorithm.  The tree generated by the RRT has branches that move in all directions.  The tree generated by the RRT* algorithm rarely has branches that move back in the direction of the parent.  The $ChooseParent()$ function ensures edges are created and always moving away from the start location.  The $Rewire()$ function changes the internal structure of the tree to ensure internal vertices do not add unnecessary steps on any discovered path.  The $ChooseParent()$ and $Rewire()$ functions guarantee the paths discovered will be asymptotically sub-optimal because these functions are always minimizing the costs to reach each node within the tree.

\section{Dynamic Replanning}

\subsection{Overview}

A real world environment is not static, and it is full of moving obstacles.  These obstacles are often moving in unpredictable directions, which makes planning tasks to avoid them difficult.  When a moving obstacle is known and is following a known trajectory, the configuration space can be modified to account for this trajectory.  When the obstacle is unknown, the robot will need to be able to dynamically determine a course of action in order to avoid a collision.  In this section a method of dynamic replanning is proposed in order to avoid a random obstacle when it is detected by the robot.

\subsection{Simulation Environment}

For the following simulations the environment remains very similar.  The robot is given a configuration space from which to build a tree using RRT* and determine the best path to reach the goal configuration from the start configuration.  In all the experiments below, the robot was allowed a tree of varying sizes to evaluate the performance with different node densities.  During the simulation a few random moving obstacles are added to the environment, described in the next section.  These obstacles represent a region of the configuration space that would be a collision if the robot were to enter.    

\subsubsection{Path Execution}

After the initial planning process, the robot begins to execute the optimal path found by the search tree.  The robot traverses the optimal path by selecting the next node and required velocity vector to reach it, see lines 3 and 4 of Algorithm 6.  This process is described in Algorithm 7 below.  When the vertex is reached the robot changes the velocity vector to move toward the next node.  This process continues until the robot reaches the goal node.  If the robot encounters a random moving obstacle that is obstructing the path a replan event occurs.

\begin{figure}[!t]
\removelatexerror
\begin{algorithm}[H]
\DontPrintSemicolon
\caption{ExecutePath()}
SetObsDestination($numObs$)\;
SetObsVelocities($numObs$)\;
SetRobotDestination()\;
SetRobotVelocity()\;

 \While {$robotLocation != GOAL$}{
	UpdateObsLocation($numObs$)\;
	UpdateRobotLocation()\;
	\If{$Replan$}{
		DoReplan()\;
	}
} 

\end{algorithm}
\vspace{-10pt}
\end{figure}

\subsection{Random Moving Obstacles}

The random moving obstacles force the robot to dynamically plan around the obstacle using RRT*.  In order for the obstacles to move about the environment, a graph is created to provide the paths between the static obstacles, and the vertices are the intersections of these paths.  Upon initialization of the simulation the obstacles are placed at random vertices.  The vertices are chosen such that the robot will have a chance to move before encountering a random obstacle.  When the simulation begins the moving obstacles choose a random adjacent vertex and begins moving toward that vertex, see lines 1 and 2 of Algorithm 6.  When the vertex is reached a new random vertex is chosen and the obstacle moves in the new direction, line 6 of Algorithm 6.  

\subsubsection{Random Obstacle Detection}

Robots operating in a real world scenario will have sensors, such as a LIDAR, to detect both static and dynamic obstacles.  Sensors are not included in this simulation.  Instead a detection range is placed on the robot.  The simulation controls whether or not a moving obstacle is within the detection range of the robot (lines 7 and 8 of Algorithm 7).  If a moving obstacle is within range the $Steer()$ function is used, by the simulation, to determine if any static obstacles are blocking the robot's line of sight to the moving obstacle.  

\begin{figure}[!t]
\removelatexerror
\begin{algorithm}[H]
\DontPrintSemicolon
\caption{UpdateRobotLocation()}
$robotLocation \leftarrow robotLocation + robotVelocity$\;
\If{$robotLocation$ == $robotDestination$}{
	$robotDestination \leftarrow $GetNextPathLocation()\;
	SetRobotVelocity()\;
}
\While {$obsIndex$ $<$ $numObs$}{
	$obsDistance \leftarrow $GetDistance($robotLocation$,...\;
				\Indp \Indp ObsLocation($obsIndex$))\; 
\Indm \Indm
	\If {$obsDistance$ $<$ $robotRange$}{
		$obs_{path} \leftarrow  $Steer($robotLocation$, $obsLocation$)\;
		\If{ObstacleFree($obs_{path}$)}{
			\If{$IsPathBlocked($obsIndex)}{
				$Replan \leftarrow $TRUE\;
			}
			SetObsVisible($obsIndex$)\;
		}
	}
} 
\end{algorithm}
\vspace{-10pt}
\end{figure}

The obstacle must be observed for a minimum of two time steps in order to determine the direction that the obstacle is moving.  Once the direction is observed the robot can determine if the moving obstacle is blocking the path or not, line 11 of Algorithm 7.  If the robot decides that the path is blocked, the replanning event begins.

\subsection{Path Replanning}

Path replanning begins with the determination of whether or not the moving obstacle is blocking the path, described in the next section.  Algorithm 8, below, lists all the steps executed during the replanning process.  The next step is to find the location along the optimal path that is beyond the obstacle.  Next, the tree generated by RRT* is modified and expanded in order to find a path around the obstacle.  Finally, the best path around the obstacle is chosen and the execution of this sub-path begins.  Each of these steps is described in the following sub-sections.

\begin{figure}[!t]
\removelatexerror
\begin{algorithm}[H]
\DontPrintSemicolon
\caption{$T \leftarrow $DoReplan()}
InvalidateNodes()\;
GetReplanGoalLocation()\;
SetReplanSamplingLimits() \;
Rewire($T$, $Q_{all}$, $NULL$, $q_{robot}$)\;
RRT*($q_{robot}$)\;
SetReplanPath()\; 
\end{algorithm}
\vspace{-10pt}
\end{figure}

\subsubsection{Path Obstruction}

The method for determining if the moving obstacle is blocking the path is a series of trigonometric functions using a direction vector from the robot to the moving obstacle and a comparison between the robot velocity vector and the moving obstacle velocity vector.  Since the configuration space is 2-Dimensional, the inverse tangent can be used to find the angles of the vectors.    To obtain the direction vector to the moving obstacle the equation is: 
\begin{equation}
angle_{direction} = atan2((Y_{obs} - Y_{robot}), (X_{obs} - X_{robot})).
\end{equation}  
Where \( (X_{robot},Y_{robot})\) is the position of the robot and \( (X_{obs},Y_{obs})\) is the position of the obstacle.  This will return an angle in degrees over the range \( (-180, 180)\).  Similarly the angle of the robot's velocity vector can be obtained using the following equation:  
\begin{equation}
angle_{V_{robot}} = atan2(V_{j}, V_{i}).
\end{equation} 
Where $V_{i}$ and $V_{j}$ are the X and Y components of the robot velocity vector.  Using the angles from (1) and (2) the difference can be taken to see if they are similar.  If the absolute value of the difference between the two angles is less than some threshold, then the robot is moving toward the moving obstacle.  Note, the angle difference is normalized to be in the range\( (-180, 180)\) before the absolute value is taken.  This is done for all angle comparisons:  
\begin{equation}
|angle_{direction} - angle_{V_{robot}}| < angle_{thresh}.
\end{equation} 
If the robot is moving in the direction of the random obstacle the velocity vectors are examined.  Substituting the obstacle velocity into (2) above the angle of the obstacle velocity can be obtained.  Next, the differences between the velocity vectors is found:
\begin{equation}
angle_{V_{diff}} = |angle_{V_{robot}} - angle_{V_{obs}}| .
\end{equation}
There are three possibilities from this point.  If the angle difference between the velocity vectors is less than the angle threshold, then the robot and the obstacle are moving in a similar direction.  Second, if the angle difference between the velocity vectors is greather than \( 180 - angle_{thresh}\), then the robot and the obstacle are approximately moving toward each other.  Last, if the angle difference falls outside of these ranges the moving obstacle and the robot are moving in different directions.  

For the first case:  The robot will simply follow the obstacle, until the obstacle changes direction, or the path takes the robot away from the obstacle.  The robot will then choose from one of the other conditions.  For the second case:  The robot quickly activates a replan event to get out of the way.  The random obstacle may move out of the way on its own, but there is no way of predicting that will occur.  Finally, if the robot and the obstacle are moving in different directions, the robot ignores the obstacle unless it gets too close.  This third condition catches the event that the robot moves out from a corner and a random obstacle is detected very close by.  This event is best summed up with the following example:  When two people approach a hallway intersection they will run into each other if they continue on their current course, it is only when they see each other that they can adjust to avoid a collision.  

\subsubsection{Select Replan Goal Location}

The second step in the replan event, line 2 in Algorithm 8, is to find a location that will navigate the robot around the random obstacle.  First, any nodes that are in collision with a random moving obstacle are invalidated, not deleted.  The only exception is the goal location of the optimal path.  After this step an assumption had to be made to simplify and speed up the rest of the replanning process.  The assumption is that the robot is currently following the best path in order to reach the goal location and should return to this path after the moving obstacle is avoided.  Using this assumption only the nodes along the optimal path are examined.  Nodes that are farther from the robot than the random obstacle are candidate nodes.  The node on the optimal path that is immediately following the node that is closest to the obstacle, without colliding, will be the replan goal location.  

\subsubsection{Modify Search Tree}

The third step is to modify the original search tree in order to find a way around the moving obstacle.  First a node is added to the tree at the robot's current location.  Using the distance to the replan goal node as a metric, a sampling area is established, line 3 in Algorithm 8.  Then, using $Rewire()$, every node within the sampling area is rewired such that the robot's current location becomes the parent of that node.  New nodes are then sampled within this area and added to the tree using RRT*.  Since there are already many nodes in the tree only a small number will need to be added.  However, the number of nearest neighbors used during the $ChooseParent()$ function and the $Rewire()$ function is increased.  This increase allows each new node to direct the existing tree toward the replan goal location. 

\subsubsection{Sub-path Selection and Execution}

When the search tree modification is complete the best path to the replan goal location is found.  Path execution will begin again as it did at the beginning of the simulation.  When the robot reaches the replan goal location, the execution of the original optimal path resumes.  If the robot encounters another moving obstacle and determines the path is obstructed again.  The replanning is repeated, however the replan goal location will always be a node on the original optimal path.

\section{Results}

\subsection{The Simulation Environment}

The environment for all of the experiments is a complex 2-Dimensional environment that will also serve as the configuration space for the robot.  The environment is complex due to the number of obstacles and several narrow passages.  There are also several sub-optimal paths where the algorithm may get stuck.  For each experiment the path cost is measured in Euclidean distance.  The environment is also intended to mimic a potential real world situation where there would be streets or sidewalks and open areas such as parks and plazas.   In the RRT* results presented below, the algorithm is allowed a maximum of 5000 nodes.  The optimal path length is 98.48 units.

\subsection{RRT* Results}

The RRT* evaluation was conducted in the following way:  There is not a growth factor for extending the tree and the tree is goal oriented.  The expecation as the tree grows will be long branches.  These branches are often inefficient, the RRT* $Rewire()$ function will remove these long inefficient branches as the algorithm executes in favor of shorter, lower cost branches.  The algorithm also has a maximum number of nearest neighbors, or neighborhood size that is configurable to the algorithm.  This implementation of RRT* has a maximum number of nearest neighbors equal to 1\% of the total number of nodes.  
 
Fig.~\ref{F.movietree2:c} shows the result of the search tree using the RRT* algorithm.  The best path length found is 103.96 units.

\subsection{Dynamic Replanning Results}

The simulation results shown below demonstrate the robot's ability to plan a path around the moving obstacle and reach the goal.  Using RRT* during the replanning step allows an efficient path to be found to avoid the moving obstacle and continue on the original optimal path.  Since the obstacles move randomly, it is possible for the robot to execute the optimal path and never be obstucted by a moving obstacle.  Only examples where the robot did encounter these obstacles are shown.

\begin{figure}
  \begin{minipage}{.5\columnwidth}
    \includegraphics[width=\columnwidth,height=5cm]{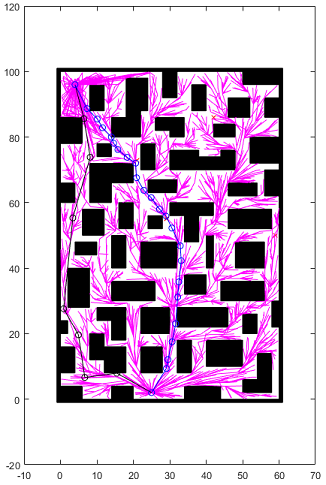}
    \subcaption{Search tree and robot path.}
    \label{F.movietree1:a}
  \end{minipage}%
  \begin{minipage}{.5\columnwidth}
    \includegraphics[width=\columnwidth,height=5cm]{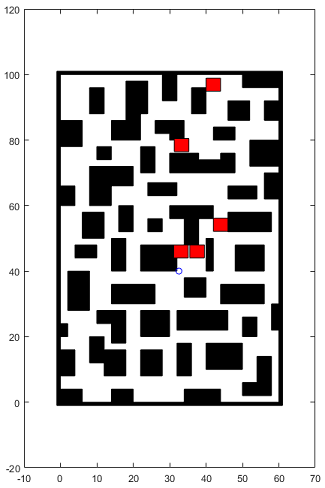}
    \subcaption{Moving obstacles block the path.}
    \label{F.movietree1:b}
  \end{minipage}%

\medskip

  \begin{minipage}{.5\columnwidth}
    \includegraphics[width=\columnwidth,height=5cm]{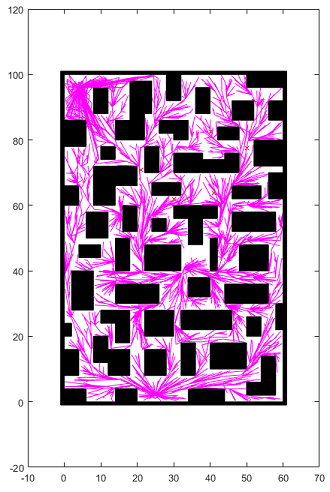}
    \subcaption{Search tree after modification during replanning.}
    \label{F.movietree1:c}
  \end{minipage}%
  \begin{minipage}{.5\columnwidth}
    \includegraphics[width=\columnwidth,height=5cm]{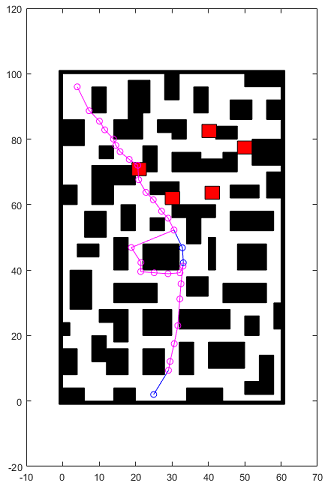}
    \subcaption{Final obstacle positions and the executed path.}
\label{F.movietree1:d}
  \end{minipage}
  \caption{Replanning results from the first set}
\label{F.movietree1}
\vspace{-10pt}
\end{figure}

The first set of results use a search tree containing 2000 nodes.  Upon completion of the search tree the moving obstacles are placed randomly within the configuration space, and the simulation begins.  Fig.~\ref{F.movietree1:a} shows the search tree found by the robot upon reaching 2000 nodes.

When executing this path the robot encounters two random obstacles near the center of the configuration space.  One obstacle moves across the path and obstructs the robot.  The robot triggers a replanning event at this time.  Fig.~\ref{F.movietree1:b} shows when the robot encountered the moving obstacles and replanned to avoid them.  A second obstacle is nearby and can be seen by the robot and must be considered when replanning.  Following the replanning steps in Algorithm 8, the robot will select a goal location, then modify the search tree to avoid the obstruction.

Fig.~\ref{F.movietree1:c} shows the full modified search tree.  There is an obvious empty region of the configuration space where the moving obstacles are located.  The abscence of branches within this area shows the robot has considered this space as obstacle space, rather than free space.  Note, the nodes in this region are not removed, they are considered invalid.  If the robot should need to replan again, and this area is free of any moving obstacles, these nodes would be available.

When the robot reaches the replanning goal location the original path can resume.  Fig.~\ref{F.movietree1:d} shows the completed path, along with the final positions of the random moving obstacles.  The magenta line shows the path followed by the robot.  The sections in blue are unexecuted portions of the original path.

\begin{figure}
    \begin{minipage}{.5\columnwidth}
    \includegraphics[width=\columnwidth,height=5cm]{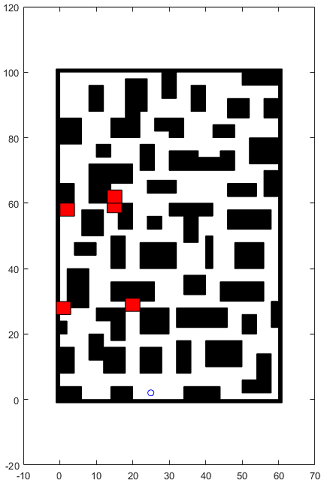}
    \subcaption{Initial positions of the moving obstacles.}
    \label{F.movietree2:a}
  \end{minipage}%
   \begin{minipage}{.5\columnwidth}
    \includegraphics[width=\columnwidth,height=5cm]{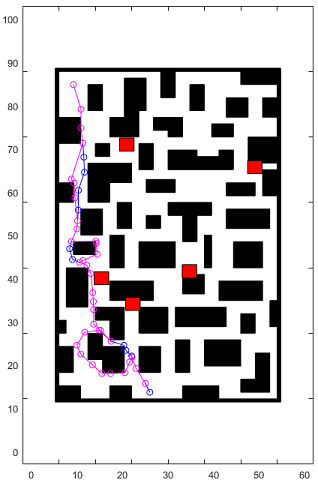}
    \subcaption{Final obstacle positions and the executed path.}
    \label{F.movietree2:b}
  \end{minipage}%

\medskip

   \begin{minipage}{.5\columnwidth}
    \includegraphics[width=\columnwidth,height=5cm]{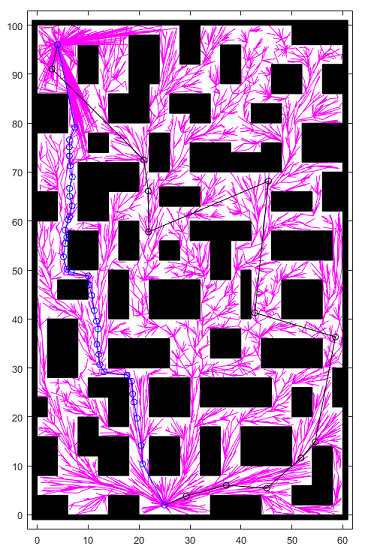}
    \subcaption{RRT* with 5000 nodes.}
    \label{F.movietree2:c}
  \end{minipage}%
    \begin{minipage}{.5\columnwidth}
    \includegraphics[width=\columnwidth,height=5cm]{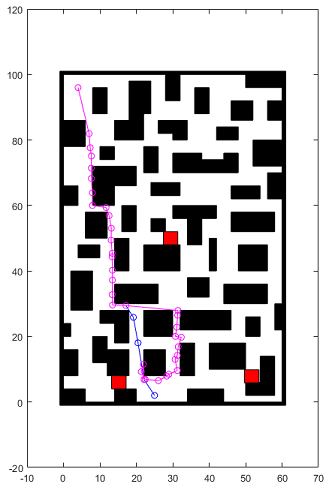}
    \subcaption{Final obstacle positions and the executed path.}
    \label{F.movietree2:d}
  \end{minipage}
  \caption{A sample RRT* search tree containing 5000 nodes and replanning results from the second and third sets.  }
\label{F.movietree2}
\vspace{-10pt}
\end{figure}

The second set of results is similar to the first, with one exception.  The random moving obstacle initial positions were selected in order to increase the probability the robot would encounter one, or many, while executing the path.  Fig.~\ref{F.movietree2:a} shows the starting locations of the moving obstacles.  The robot was obstructed three times during the execution of the path and successfully planned around the moving obstacle each time.  Fig.~\ref{F.movietree2:b} shows the final positions of moving obstacles and the path followed by the robot.

The final set of results is a simulation with a search tree containing 5000 nodes and 3 obstacles moving at random in the configuration space.  The three moving obstacles in this simulation were placed similarly to those in the second simulation.  In this simulation the robot encounters a moving obstacle very early in the execution of the path.  The robot finds a path around the moving obstacle and completes the path.  Fig.~\ref{F.movietree2:d} shows the final positions of the obstacles and the executed path.

\section{Conclusion and Future Work}

The replanning method presented performs well and is a good first step toward a more robust method of replanning when unknown randomly moving obstacles obstruct the robot's path.  Future work will research a floating replan goal location to minimize the total remaining cost to the query goal.  Minimizing the modifications to the original search tree is another area of improvement.  This method has only been implemented in a 2-Dimensional configuration space.  The algorithm must be expanded and modified to operate in higher dimension configuration spaces.  The simulations up to this point have been with small numbers of moving obstacles, further research is needed to determine how the algorithm performs when there are many random moving obstacles.
Additional research is pursuing multi-robot systems.  Efficient path planning and replanning will benefit: cooperative sensing systems \cite{La_2,La_4}, formation control systems\cite{La_1, La_3, La_5, La_7, La_ICRA10} , and target tracking and observation \cite{La_6}.

\bibliography{Thesis_bibliography}
\end{document}